\documentclass[10pt,twocolumn,letterpaper]{article}

\usepackage{iccv}
\usepackage{times}
\usepackage{epsfig}
\usepackage{graphics} 
\usepackage{caption}
\usepackage{subcaption}
\usepackage{graphicx}
\usepackage{amsmath}
\usepackage{amssymb}
\usepackage{url}
\usepackage{multirow}
\usepackage{algorithm}
\usepackage[noend]{algpseudocode}
\usepackage{appendix}
\usepackage{subfiles}
\usepackage{array, makecell}
\usepackage[accsupp]{axessibility} 

\usepackage[pagebackref=true,breaklinks=true,letterpaper=true,colorlinks,bookmarks=false]{hyperref}

\iccvfinalcopy 

\def\iccvPaperID{10} 

\ificcvfinal\fi

\begin{document}

\title{Iterative Distillation for Better Uncertainty Estimates in  Multitask Emotion Recognition}

\author{\parbox{16cm}{\centering
    {\large Didan Deng$^1$, Liang Wu$^2$, Bertram E. Shi$^3$}\\
    {\normalsize
    Department of Electronic and Computer Engineering, \\Hong Kong University of Science and Technology, Kowloon, Hong Kong\\
    \{ddeng$^1$, lwuat$^2$\}@connect.ust.hk\qquad\quad eebert@ust.hk$^3$}}
}
\maketitle
\ificcvfinal\thispagestyle{empty}\fi

\begin{abstract}
    When recognizing emotions, subtle nuances in displays of emotion generate ambiguity or uncertainty in emotion perception. Emotion uncertainty has been previously interpreted as inter-rater disagreement among multiple annotators. In this paper, we consider a more common and challenging scenario: modeling emotion uncertainty when only single emotion labels are available. From a Bayesian perspective, we propose to use deep ensembles to capture uncertainty for multiple emotion descriptors, \ie, action units, discrete expression labels and continuous descriptors. We further apply iterative self-distillation. Iterative distillation over multiple generations significantly improves performance in both emotion recognition and uncertainty estimation. Our method generates single student models that provide accurate estimates of uncertainty for in-domain samples and a student ensemble that can detect out-of-domain samples. Our experiments on emotion recognition and uncertainty estimation using the Aff-wild2 dataset demonstrate that our algorithm gives more reliable uncertainty estimates than both Temperature Scaling and Monte Carol Dropout.

\end{abstract}

\section{Introduction}

Understanding human affective states is an essential task for many interactive systems (\eg, social robots) or data mining systems (\eg, user profiling). However, unlike object recognition tasks, emotion perception is strongly affected by personal bias, cultural backgrounds and contextual information (\eg, environment), which increases the uncertainty of emotion perception. 

To obtain a gold standard for emotion recognition, it is common to invite a number of annotators and take the most-agreed emotions as hard labels in emotion datasets \cite{mavadati2013disfa,valstar2015fera,zhang2018facial,zafeiriou2017aff}. In datasets with huge number of samples \cite{mollahosseini2017affectnet}, it is expensive to invite many annotators. Therefore, each sample is often annotated by one expert only. Single emotion labels cannot capture inter-rater disagreement. 
\begin{figure}
    \centering
    \includegraphics[width=1.0\columnwidth]{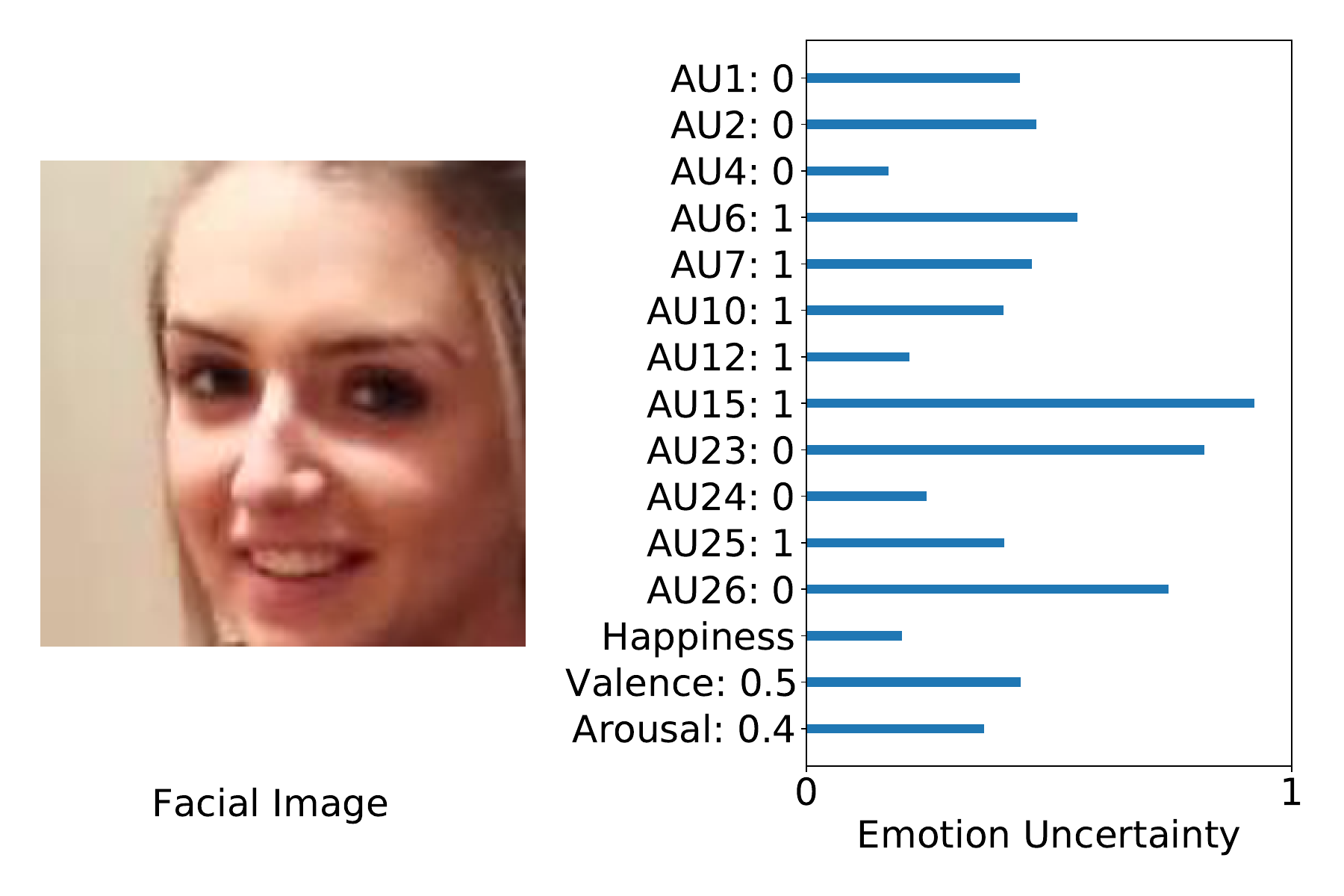}
    \caption{An example facial image with classifications generated by our model. The classifications are shown in the labels in the left. The estimated uncertainty is shown in the bar graph in the right. For example, the emotion is classified as "Happiness". The estimated valence and arousal scores are 0.5 and 0.4. The presence or absence of each AU is indicated by 1 or 0. The emotion uncertainty is the normalized Shannon's entropy (\ie, divided by its information length). Large uncertainty indicates low confidence. For example, AU15 "lip corner depressor" is indicated as present, but with high uncertainty (low confidence). It is not present in the example image. }
    \label{fig:teaser}
\end{figure}

Previous work often related emotion ambiguity (uncertainty) with the variability among multiple raters' annotations. For example, Mower \etal \cite{mower2009interpreting} characterized ambiguity using probability distributions assigned to the emotion classes. Han \etal \cite{han2017hard} defined emotion uncertainty as \textit{perception uncertainty} (\ie, inter-rater disagreement). The solutions cannot be applied to emotion datasets with only single labels. 

To address this problem, we adopt the Bayesian viewpoint and interpret emotion uncertainty as uncertainty in the posterior distribution over model weights. It is affected by the nosie, data distribution, and the model we choose.  It does not require multiple raters' annotations, as required by the \textit{perception uncertainty}. 

In addition, we consider uncertainty simultaneously for multiple types of emotion labels (\ie, facial action units, basic emotions, valence and arousal), whereas past studies considered only single label types. Our intuition is that human affective states are quite complex, and should be described using a comprehensive set of emotional descriptors. Uncertainty among different emotion labels may be correlated. For example, the relationship between valence and arousal may be related to the uncertainty in perceived valence \cite{brainerd2018emotional}.  

The recently released Aff-wild2 \cite{kollias2018aff, kollias2021analysing} facilitates multitask emotion solutions \cite{kollias2021distribution,kollias2021affect, deng2020multitask}. The Aff-wild2 dataset has three types of emotion labels: facial action units, emotion categories, valence and arousal. Past emotion datasets \cite{mavadati2013disfa,valstar2015fera,zhang2018facial,kollias2019deep} usually have one or two types of emotion labels. However, Aff-wild2 dataset only provides single emotion labels, not multiple annotators' labels. 

Using data from the Aff-wild2 dataset, we train deep ensembles with self-distillation algorithm to improve emotion recognition and uncertainty estimation. The obtained networks produce both emotions labels and the estimated uncertainty. The uncertainty is measured by Shannon's entropy computed over the probabilistic output. We give an example in Figure \ref{fig:teaser}, showing the outputs of our model given a facial image input.

Our primary contributions are as follows:

\begin{itemize}
    \item For better uncertainty estimation performance, we propose to apply deep ensembles learned by multi-generational self-distillation. The iterative training of neural networks improves not only uncertainty estimation, but also multitask emotion recognition.
    \item We design Efficient Multitask Emotion Networks (EMENet) for video emotion recognition. The visual model (EMENet-V) only has $1.68M$ parameters. The visual-audio (EMENet-VA) model has $1.91M$ parameters. 
    \item We show that single models can estimate uncertainty reliably on in-domain data, and that the ensembles can detect out-of-distribution (OOD) samples.
\end{itemize}

\section{Related Works}
\subsection{Uncertainty in Emotion}
In emotion recognition, uncertainty often refers to \textit{perception uncertainty}, in other words, inter-rater disagreement, requiring multiple annotators. Han \etal \cite{han2017hard} took the standard deviation of $K$ emotion labels given by $K$ annotators as perception uncertainty. Zhang \etal \cite{zhang2014agreement} used Kappa coefficient to represent inter-rater agreement level. Uncertainty in emotion recognition has also been used to refer to the uncertainty in probabilistic models. A work in speech emotion recognition used a probabilistic Gaussian Mixture Regression (GMR) model to get the uncertainty of samples \cite{dang2017investigation}. The authors found the emotion model performs better in low-uncertainty regions than high-uncertainty regions. Dang \etal \cite{dang2017investigating} also used probabilistic models, and applied uncertainty when fusing predictions from sub-systems of multiple modalities. These past methods relied on hand-crafted features. 

\subsection{Uncertainty Estimation}

Ensemble-based methods are alternatives to Bayesian methods for estimating decision uncertainty. A Deep Ensemble \cite{lakshminarayanan2016simple} consists of several neural networks with the same architecture, but their weights are initialized independently. From a Bayesian viewpoint, the learned weights are "sampled" from a posterior distribution. Deep ensembles have been shown to provide uncertainty estimates robust to dataset shifts \cite{ovadia2019can}. Similar to deep ensembles, the Monte Carol Dropout (MC Dropout) \cite{gal2016dropout} is a Bayesian approximation method for estimating uncertainty. MC Dropout method uses dropout during both training and testing. During inference, a dropout model is sampled $T$ times, and the $T$ predictions are averaged. Temperature Scaling \cite{guo2017calibration} (TS) is a post-hoc calibration method to improve uncertain estimation. It optimizes the temperature value of the softmax function on a held-out validation set. The advantage of TS is that it does not increase computation during inference, but it is prone to overfitting.

\subsection{Knowledge Distillation}
Knowledge Distillation~\cite{hinton2015distilling} was firstly proposed by Hinton \etal for model compression. A special case of knowledge distillation is the self-distillation algorithm \cite{furlanello2018born}, where the student model has the same architecture as its teacher model. The student model usually outperforms its teacher model, as shown in \cite{furlanello2018born, xie2020self}. The multi-generational self-distillation algorithm uses the student model in the previous generation as the teacher model in the next generation. As the number of generations increases, generalization performance improves \cite{mobahi2020self}.  Some studies have studied the reasons behind this phenomenon. For example, Mobahi \etal \cite{mobahi2020self} proved mathematically that self-distillation amplifies regularization in the Hilbert space. Zhang \etal \cite{zhang2020self} related self-distillation to label smoothing, a commonly-used technique to prevent models from being over-confident \cite{muller2019does}.  They suggested that the regularization effect of self-distillation results from instance-level label smoothing. In this work, we aim to investigate the self-distillation for improving uncertainty performance by extending single models to deep ensembles. 

\section{Methodology}
 
\subsection{Notations}
We denote the training set as $\{X, Y\}$, where $X$ denotes the input data and $Y$ denotes the ground truth labels. The input data can be divided into two categories $\{X^{vis}, X^{aud}\}$. $X^{vis}$ represents the visual data and $X^{aud}$ represents the audio data. $X^{vis}$ contains facial images, where $X^{vis}=\{x_i^{vis}| x_i^{vis}\in\mathbb{R}^{3\times H\times H}\}_{i=1}^{N}$. The facial images are RGB images with a height (width) of $H$ pixels. $X^{aud}$ contains mel spectrograms: $X^{aud} = \{x_i^{aud}|x_i^{aud}\in\mathbb{R}^{W\times W}\}_{i=1}^N$. The mel spectrograms have two dimensions: the number of mel-filterbank features and the number of audio frames. They are both $W$ in our experiments.

The ground labels $Y$ can be divided into three types: $Y = \{Y^{AU}\in\mathbb{R}^{N\times 12}, Y^{EXPR}\in\mathbb{R}^{N\times 7}, Y^{VA}\in \mathbb{R}^{N\times 2}\}$. $Y^{AU}$ contains 12 facial action units labels, including AU1, AU2, AU4, AU6, AU7, AU10, AU12, AU15, AU23, AU24, AU25 and AU26. They are multi-label binary values, denoting the presence or absence of corresponding action unit. $Y^{EXPR}$ are one-hot vectors denoting 7 basic emotions: neutral, anger, disgust, fear, happiness, sadness and surprise. $Y^{VA}$ are given by continuous values representing valence and arousal in range $\{-1, 1\}$. In our experiments, we transform regression tasks into classification tasks by discretizing continuous values. We discretize the valence score or the arousal scores into 20 bins, so that the shape of $Y^{VA}$ changes to $N\times 40$.

The single model function is denoted by $f_\theta$, where $\theta$ denotes the parameters. The ensemble model with $T$ models is denoted by $F_T$, where $F_T(x) = \frac{1}{T}\sum_{t=1}^T \sigma\left(f_{\theta_t} (x)\right)$. $\sigma(\cdot)$ is an activation function. The output of the ensemble is the average of its members' outputs. 

In our teacher-student algorithm, a teacher ensemble with $T$ models can be denoted as $F_T^{tea}$. The soft labels generated by the teacher ensemble are denote as $ F_T^{tea}(X)$. The student ensemble in the $k^{th}$ generation is denoted as $F_T^{stu_k}$. The soft labels generated by this ensemble for the $k+1$ generation is $F_T^{stu_k}(X)$. We run the self-distillation algorithm for $K$ generations in total.

\subsection{Architectures}
\begin{figure}
    \centering
    \includegraphics[width=1\columnwidth]{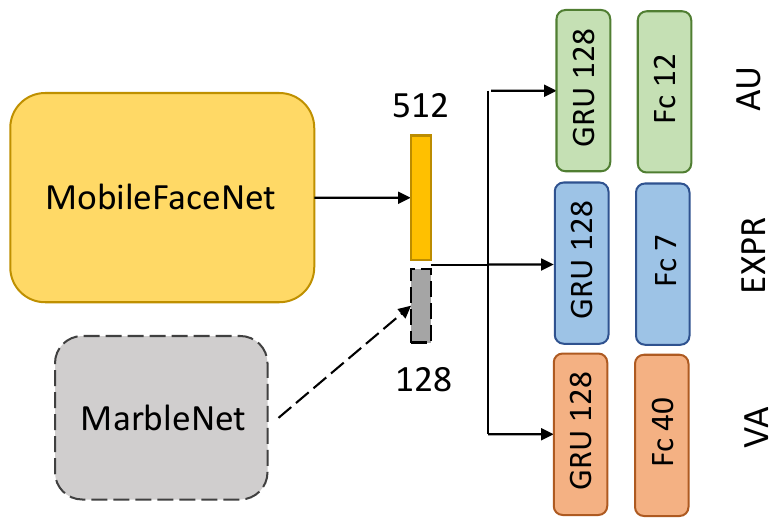}
    \caption{Our efficient model architecture. For visual modality, the feature extractor is the MobileFaceNet, and visual feature vector is a 512-dimensional vector. The weights surrounded by dashed curves are only included in the multimodal model architecture. For mutlimodal model, the MarbleNet is the audio feature extractor, and the audio feature vector is a 128-dimensional vector. The visual feature vector and the audio feature vector are concatenated before they are fed into temporal models. }
    \label{fig:arch}
\end{figure}

We aim to design efficient architectures while maintaining high performance in emotion recognition. Previous studies \cite{deng2020multitask,deng2020mimamo,kollias2019expression, kollias2019face} on video emotion recognition showed that CNN-RNN architectures generally outperformed CNN architectures. This indicates that affective states have strong temporal dependencies. Therefore, we choose efficient CNN architectures as feature extractors, and use GRU layers as temporal models to integrate information over time.

For the visual modality, our model receives a sequence of facial images as input. The facial images are firstly processed by the MobileFaceNet \cite{chen2018mobilefacenets}. The MobileFaceNet is a light-weighed CNN originally designed for face recognition on mobile devices. The model was pretrained on face alignment task \cite{chenface2021}. We then finetuned the pretrained CNN weights. The feature vector is a 512-dimensional vector for each input image. The detailed architecture of the MobileFaceNet is given in \cite{chen2018mobilefacenets}. 



For the visual-audio model, we used a 1D-CNN to extract audio features from mel spectrograms. The audio CNN (MarbleNet) was proposed by Jia \etal \cite{jia2021marblenet} for voice activity detection. It has only $88K$ parameters. The audio feature vector extracted from one mel spectrogram is a 128-dimensional vector. The input is a $64\times64$ mel spectrogram, which is produced by extracting 64 mel-filterbank features from 64 audio frames (640ms). In our experiments, we sample a sequence of facial images as well as a sequence of mel spectrograms. The sample rate is same as the frame rate of the input video file. We denote the sequence length by $L$ and the batch size by $B$. The shapes of inputs to our visual-audio model are $(B, L, 3, 112, 112)$ for $X^{vis}$ and $(B, L, 64, 64)$ for $X^{aud}$. After the inputs are processed by feature extractors, the visual features and audio features are concatenated. This results in $(B, L, 640)$ feature vectors that are fed into the temporal models. 

Each task has their own temporal model, which consists of one GRU layer, a ReLU activation function, and a linear output layer. The hidden sizes of all GRU layers are 128. We apply a $50\%$ random dropout on the input features to the temporal models. For the final activation function, we use Sigmoid for the AU detection, and Softmax for 7 basic emotions, and valence/arousal prediction.



\subsection{Loss Functions}
To train teacher models, we minimize the loss functions between the teacher outputs and the ground labels. We refer to these losses as \textit{supervision losses}. To train student models, we minimize the loss functions between the student outputs and the soft labels, which are generated by the teacher models or the student models in the previous generation. We refer to these loss functions as the \textit{distillation losses}.

\textbf{\textit{Supervision Losses}}. For facial action units detection, we use a sum of class-reweighted binary cross entropy (BCE) functions. We reweight the losses using $p_c$ based on the ratios between positive samples and negative samples in the training data. 

\begin{align}
    \mathcal{L}^{AU} (y, \Tilde{y}) = \frac{1}{C}\sum_{c=1}^C \mathbf{BCE}(y_c, \Tilde{y}_c), \\\label{eq:bce_super}
    \mathbf{BCE}(y_c, \Tilde{y}_c) = - [ p_cy_c\cdot\log\left( \sigma(\Tilde{y}_c)\right) + \\\nonumber  (1-y_c)\cdot\log\left(1-\sigma(\Tilde{y}_c)\right)], \\ 
    p_c= \frac{\#\: negative\: samples\: in\: class\: c}{\#\: positive\: samples\: in\: class\: c}.
\end{align}

Variables $y$ denote ground truth labels and $\Tilde{y}$ denote inferences generated by the teacher model. $\sigma(\cdot)$ in Equation \ref{eq:bce_super} denotes the Sigmoid function. $C$ is the total number of action units.

For basic emotion categories, we use a reweighted cross entropy (CE) function. The weights are determined by the distribution of different classes in the training set.
\begin{align}
    \mathcal{L}^{EXPR}(y, \Tilde{y}) = \mathbf{CE}(y, \Tilde{y}), \\
    \mathbf{CE}(y, \Tilde{y}) = -p_c\sum_{c=1}^C y_c \cdot\log(\Tilde{y}_c) \label{eq: ce_expr}.
\end{align}

For valence/arousal predictions, we use the Concordance Correlation Coefficient (CCC) between the scalar outputs and the ground truth labels. The CCC is defined as follows:

\begin{equation}
\mathbf{CCC} (y_c, \Tilde{y}_c) = \frac{2\rho \sigma_y \sigma_{\Tilde{y}}}{\sigma_y^2 + \sigma_{\Tilde{y}}^2 + (\mu_y - \mu_{\Tilde{y}})^2},
\label{eq:CCC}
\end{equation}
where $y_c$ denotes ground truth labels in a batch, and $\Tilde{y}_c$ denotes the scalar predictions of valence or arousal. $\rho$ is the correlation coefficient between the ground truth labels and the predictions. $\mu_y$, $\mu_{\Tilde{y}}$, $\sigma_y$ and $\sigma_{\Tilde{y}}$ are the means and standard deviations computed over the batch. Since our model produces a 20-dimensional softmax vector for valence/arousal, we compute the expectation values over the 20 bins in the range of $[-1, 1]$ to transform probabilistic outputs to scalar outpus.

We compute two CCCs: one for valence, and one for arousal. The supervision loss for valence and arousal prediction is:
\begin{equation}
    \mathcal{L}^{VA}(y, \Tilde{y}) = \sum_{c=1}^2 (1-\mathbf{CCC}(y_c, \Tilde{y}_c)).
    \label{eq:va_lsos}
\end{equation}

\begin{figure*}[hbtp]
    \centering
    \includegraphics[width=0.9\textwidth]{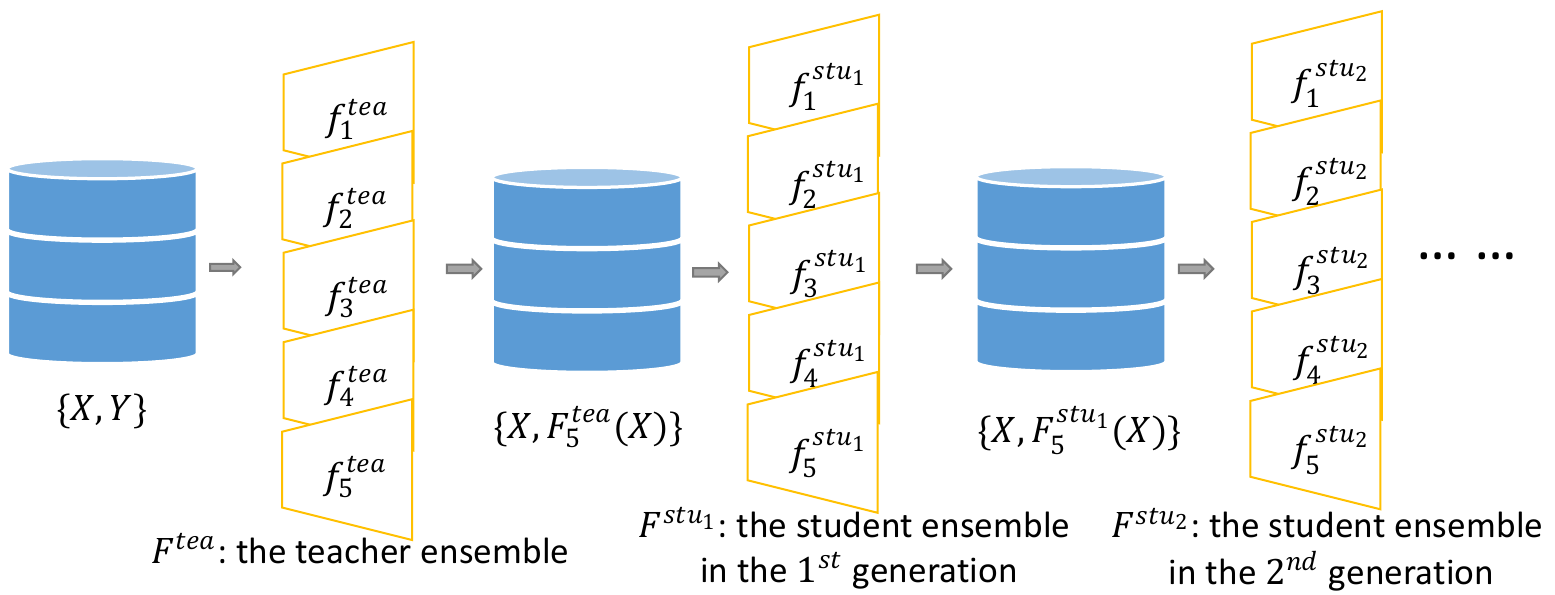}
    \caption{The diagram of our teacher-student algorithm. }
    \label{fig:diagram}
\end{figure*}

\textbf{Distillation losses}. For action units detection, we use the binary cross entropy between the soft labels and the outputs of the student models. 
\begin{align}
    \mathcal{H}^{AU}(y^{tea}, \Tilde{y}^{stu}) = \mathbf{BCE}(y^{tea}, \Tilde{y}^{stu}), \\
    \mathbf{BCE}(y^{tea}, \Tilde{y}^{stu}) = -[
    p_c y^{tea}\cdot\log\left(\sigma(\Tilde{y}^{stu})\right) + \label{eq:bce_distill}\\ \nonumber
    (1-y^{tea})\cdot\log\left(1-\sigma(\Tilde{y}^{stu})\right)], 
\end{align}
where $p_c$ in Equation \ref{eq:bce_distill} is the same as $p_c$ in Equation \ref{eq:bce_super}. $y^{tea}$ represents the soft labels. $y^{tea} = F^{tea}_T(x)$ if it is in the first generation. $y^{tea} = F^{stu_k}_T(x)$ if it is in the $(k+1)^{th}$ generation. $\Tilde{y}^{stu}$ is the output of a single student model.

For expression recognition, the distillation loss we use is the KL divergence (KLD) loss between the soft labels and the student outputs.

\begin{equation}
    \mathcal{H}^{EXPR}(y^{tea}, \Tilde{y}^{stu}) = \mathbf{KLD}(y^{tea}, \Tilde{y}^{stu}).
\end{equation}

For valence and arousal prediction, we still use negative CCC loss between the soft labels and scalar outputs of the student model.

\begin{equation}
    \mathcal{H}^{VA}(y^{tea}, \Tilde{y}^{stu}) = \sum_{c=1}^2 (1-\mathbf{CCC}(y^{tea}, \Tilde{y}^{stu}))
    \label{eq:va_distill},
\end{equation}
where $ y^{tea}$ and $ \Tilde{y}^{stu}$ are the scalar outputs of the teacher ensemble and the student model.

\textbf{Combination of losses}. We take a weighted sum of the losses for different tasks. For example, when training teacher models, we use a combination of supervision losses for different emotion tasks:
\begin{equation}
    \mathcal{L} = \lambda^{AU}\mathcal{L}^{AU} + \lambda^{EXPR}\mathcal{L}^{EXPR} + \lambda^{VA}\mathcal{L}^{VA}.
    \label{eq: teacher}
\end{equation}

When training student models, we use a combination of distillation losses for all tasks.
\begin{equation}
    \mathcal{H} = \lambda^{AU}\mathcal{H}^{AU} + \lambda^{EXPR}\mathcal{H}^{EXPR} + \lambda^{VA}\mathcal{H}^{VA}.
    \label{eq: student}
\end{equation}

When training multiple tasks, it is important to balance the weights of different losses according to the difficulty levels of different tasks. We propose a heuristic method to balance the weights. The weight of the $i^{th}$ task's loss depends on the number of epochs with no performance improvement on the validation set. If it is larger, we assign larger weights to this task's loss to increase its influence on the gradients. Appendix B shows the pseudo code for this heuristic method and ablation studies.

\subsection{Algorithm}
Our algorithm is a special case of the self-distillation algorithm. In the original self-distillation algorithm \cite{furlanello2018born,mobahi2020self}, the teacher model and the student model are single models with the same architecture. In our algorithm, we propose to use deep ensembles for the following benefits:

1. The deep ensembles can be naturally trained on a distributed system. The parallel computing facilities parallel training of each local model, which saves training time.

2. The soft labels provided by the teacher ensemble contains more reliable uncertainty information than that provided by a single teacher model.

3. We can use one single model or a few models in our student ensemble to perform emotion tasks, which brings more flexibility when it comes to computation cost.

In Figure \ref{fig:diagram}, we illustrate the teacher-student algorithm for deep ensembles. $\{X, Y\}$ is the original dataset. Most of instances in $\{X, Y\}$ only have one type of emotion labels, while other two types of emotion labels are missing. The $t^{th}$ teacher model $f_t^{tea}$ learns to fill in the missing labels. In the training batch of $\{X, Y\}$, we sample an equal number of instances for three tasks,  and then compute the loss in Equation \ref{eq: teacher}. Note that $\{f_t^{tea}\}_{t=1}^T$ are all trained on the same dataset $\{X, Y\}$, but start with different random initialization. After training $\{f_t^{tea}\}_{t=1}^T$ parallelly, we take average over their predictions on the training data. This generates the soft labels for the student models in the first generation. The soft labels are denoted as $F_T^{tea}(X)$. 

In the first generation, student models $f_t^{stu_1}$ are trained on $\{X, F_T^{tea}(X)\}$. After all student models are trained, we use the student ensemble to generate the soft labels for the next generation. We iterative the teacher-student training in order to find the best number of generations. 

\section{Experiments}
\subsection{Dataset}
We only used the video data from the Aff-wild2 \cite{kollias2018aff} dataset. The Aff-wild2 dataset has three subsets, one for each emotion task. In each subset, the data distributions are quite unbalanced. Appendix A shows the data distributions for the three subsets. The data distributions determine the class weights $p_c$ in Equation \ref{eq:bce_super} and \ref{eq: ce_expr}. Appendix A also gives the choices of $p_c$.


\subsection{Hyper-parameters}
We used the Adam \cite{kingma2014adam} optimizer. The learning rate was initialized as $1e^{-3}$. For visual model training, we trained the model for 10 epochs, and decreased the learning rate by a factor of 10 after every 3 epochs. For multimodal training, we trained the models for 15 epochs and decreased the learning rate by a factor of 10 after every 4 epochs. 

\subsection{Metrics}
\textbf{Emotion metrics}. We used the same evaluation metrics as suggested in~\cite{kollias2020analysing}. For facial AU detection, the evaluation metric is $0.5 \cdot \mathrm{F1} + 0.5 \cdot \mathrm{Acc}$, where F1 denotes the unweighted F1 score for all 12 AUs, and Acc denotes the total accuracy. For expression classification, we used $0.67 \cdot \mathrm{F1}+0.33 \cdot \mathrm{Acc}$ as the metric, where F1 denotes the unweighted F1 score for 7 classes, and Acc is the total accuracy. For valence and arousal, we evaluated them with CCC.

\textbf{Uncertainty metric}. Same to \cite{lakshminarayanan2016simple}, we evaluated the in-domain uncertainty estimation performance using the negative log-likelihood (NLL) for classification tasks (\ie, EXPR and AU) and root mean square error (RMSE) for regression tasks (\ie, valence and arousal). Lower NLL or RMSE means better uncertainty estimation performance. For out-of-domain uncertainty performance, we created an OOD detection task by importing non-facial images, and evaluated the binary classification performance using ROC (receiver operating characteristic) curves and AUC (area under the ROC curve) scores.

\begin{figure}
    \centering
    \includegraphics[width=\columnwidth]{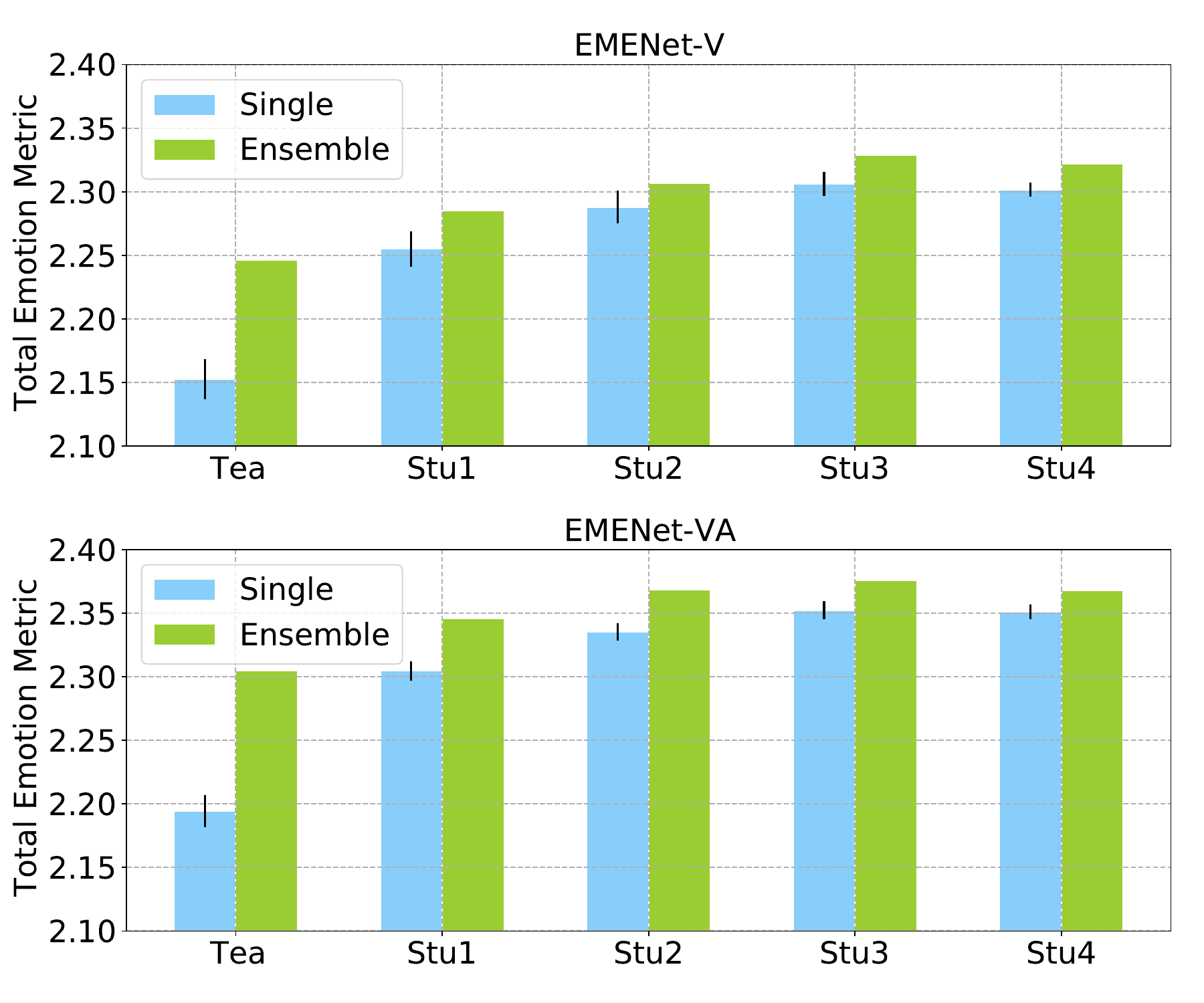}
    \caption{The total emotion metrics for the visual model (EMENet-V) and the visual-audio model (EMENet-VA), on the validation set of the Aff-wild2. For the single models' results, we average the total emotion metric over five runs, and the standard deviations are shown with error bars. "Tea" stands for the teacher model (ensemble). "Stu1" stands for the student model (ensemble) in the first generation.}
    \label{fig:total_emotion}
\end{figure}

\section{Results}

\subsection{Task Performance}
    

\begin{table}[hbtp]
    \centering
    \resizebox{\columnwidth}{!}{
    \begin{tabular}{c|c|c|c|c|c|c}
    \hline
        \multirow{2}{*}{Experiments} & \multirow{2}{*}{T}  & \multirow{2}{*}{AU} & \multirow{2}{*}{EXPR}  & \multicolumn{2}{c}{VA} & \multirow{2}{*}{Total Emotion} \\ \cline{5-6}
       & &   & &  Valence & Arousal  & \\
       \hline 
       w/o re. & 1  & 0.6773 & 0.5128 & 0.3830 & 0.5268 & 2.0999\\
        w/o re. & 5 & 0.6858 & 0.5354 & 0.4099 & 0.5537 & 2.1848 \\
       w/o re. & 10  & 0.6843 & 0.5449& 0.4105& 0.5577 & 2.1974 \\
       w/ re. &1 & 0.6632  & 0.5541 & 0.4202 & 0.5192 & 2.1527 \\
       w/ re. & 5  & 0.6808 &0.5779 & 0.4423 & 0.5455 &\textbf{2.2465}\\
       \hline
        \end{tabular}
        }
    \caption{Validation results with the teacher models using visual modality only. "w/ re." means we apply class reweighting for EXPR and AU. $T$ is the number of models in an ensemble. $T=1$ means it is a single model. Total emotion metric is the sum of all metrics of the three emotion tasks. }
    \label{tab:tea_vis} 
\end{table}

\begin{table}[hbtp]
    \centering
    \resizebox{\columnwidth}{!}{
    \begin{tabular}{c|c|c|c|c|c|c}
    \hline
        \multirow{2}{*}{Methods}  &\multirow{2}{*}{$\#$ Gen.}  & \multirow{2}{*}{$\#$ Param.} & \multirow{2}{*}{AU} & \multirow{2}{*}{EXPR}  & \multicolumn{2}{c}{VA}  \\ \cline{6-7}
       & &  & &&  Valence & Arousal \\
       \hline 
      EMENet-V &3 & 1.68M& 0.6320 & 0.4639 & 0.4942 & 0.4355 \\
      EMENet-V & 3& 8.4M & 0.6328 & 0.4704& 0.5104 & 0.4419\\
      EMENet-VA & 2& 1.91M &0.6418 &\textbf{0.5046} & \textbf{0.5355} & 0.4442 \\
      EMENet-VA & 1& 9.55M &\textbf{0.6528}  &0.5041 & 0.5326 &\textbf{0.4537} \\
       \hline
        \end{tabular}
        }
    \caption{The emotion metrics on the test set of the Aff-wild2 dataset. "$\#$ Gen." denotes the number of generations. "$\#$ Param." denotes the number of parameters. The enmsebles have five times larger number of parameters than single models.}
    \label{tab:test_res} 
\end{table}

\begin{table*}[hbtp]
    \centering
     \resizebox{\textwidth}{!}{
    \begin{tabular}{c|c|c|c|c|c|c|c|c|c|c|c|c|c}
    \hline
        Methods  & AU1 & AU2 &AU4 & AU6 & AU7 & AU10 & AU12 & AU15 & AU23 & AU24 & AU25 & AU26 & Avg. \\ 
    \hline
     Tea &0.407 & 0.348 & 0.424 & 0.435 & 0.477 & 0.437 & 0.380 & 0.389 & 0.464 & 0.459 & 0.491 & 0.442 & 0.430\\
     Stu1 &0.366 & 0.326 & \textbf{0.354} & \textbf{0.387} & 0.471 & 0.427 & 0.379 & 0.322 & 0.388 & 0.328 &\textbf{0.482} & 0.413 & 0.387\\
     Stu2 &0.358 & 0.319 & 0.355 & 0.388 & \textbf{0.468} & 0.424 & 0.383 & 0.314 & 0.367 & 0.303 & 0.487 & 0.400 & \textbf{0.381}\\
     Stu3 &0.351 & 0.312 & 0.369 & 0.395 & 0.479 & 0.435 & 0.400 & \textbf{0.310} & 0.353 & 0.287 & 0.488 & 0.398 & \textbf{0.381}\\
     Stu4 &\textbf{0.338} & \textbf{0.301} & 0.377 & 0.401 & 0.482 & 0.440 & 0.421 & 0.322 &\textbf{0.347} &\textbf{0.271} & 0.490 &\textbf{0.386} & \textbf{0.381}\\
      TS \cite{guo2017calibration} & 0.405 & 0.345 & 0.420 & 0.435 & 0.476 & 0.435 & 0.379 & 0.388 & 0.462 & 0.449 & 0.491 & 0.441 & 0.427\\
      MC \cite{gal2016dropout}& 0.408 & 0.350 & 0.455 & 0.429 & 0.472 & \textbf{0.422} &\textbf{0.362} & 0.357 & 0.402 & 0.499 & 0.485 & 0.414 & 0.421\\
       \hline
        \end{tabular}
        }
    \caption{The NLL values for 12 action units, which are evaluated on the validation set of the Aff-wild set. We compare our single teacher models and single student models with other methods, \ie, TS (temperature scaling \cite{guo2017calibration}) and MC (Monte-Carol Dropout \cite{gal2016dropout}). The model architecture used in this comparison is EMENet-V.} 
    \label{tab:AU_uncertainty} 
\end{table*}

\begin{table}[hbtp]
    \centering
    \begin{tabular}{c|c|c|c}
    \hline
        \multirow{2}{*}{Methods} & \makecell{EXPR \\ NLL}& \makecell{Valence \\ RMSE} &\makecell{Arousal \\ RMSE}   \\
        \hline
        Tea & 1.052 & 0.400 & 0.256 \\
        Stu1 & 0.911 & 0.379 & 0.234 \\
        Stu2 & \textbf{0.905} & 0.377 & \textbf{0.231}\\
        Stu3 & 0.918 & 0.373 & 0.232\\
        Stu4 & 0.957 &\textbf{ 0.370} & 0.233\\
        TS \cite{guo2017calibration} & 0.998 & - & -\\
        MC \cite{gal2016dropout} & 1.071 & 0.398 & 0.251 \\
        \hline
    \end{tabular}
    \caption{The NLL values for EXPR recognition and the RMSE values for valence and arousal prediction. Metrics are evaluated on the validation set. TS optimizes temperature for lower NLL on a held-out validation set, which is not beneficial for RMSE in regression tasks. Therefore, we only compare our models with TS for EXPR task. }
    \label{tab:expr_va}
\end{table}

\textbf{Computation cost}. We designed two model architectures for the visual modality and visual-audio modalities respectively. We refer to them as EMENet-V and EMENet-VA. The number of parameters and FLOPs for EMENet-V are $1.68M$ and $228M$. For EMENet-VA, they are $1.91M$ and $234M$. The FLOPs are the number of floating-point operations when the visual input is one RGB image (112x112) and audio input is one spectrogram (64x64).

\textbf{Class reweighting}. We show the effect of class reweighting in Table \ref{tab:tea_vis}. After applying class reweighting, we found the EXPR metric for single models was improved significantly, where the F1 score increased by $12.6\%$, and the accuracy of EXPR increased by $1.7\%$. Although the AU metric degraded after using class reweighting, its F1 score increased by $12.5\%$. The AU metric degraded because its accuracy dropped from $0.8947$ to $0.8249$. We think this is due to the highly unbalanced data distribution in the AU subset. 

\textbf{Ensemble size}. We changed the ensemble size when training teacher models without class reweighting. The results are reported in Table \ref{tab:tea_vis}. From single models ($T=1$) to ensemble models ($T=5$), the total emotion metric increased by $4\%$. However, from $T=10$ to $T=5$, the total emotion metric only increased by $0.58\%$. We kept the ensemble size $T=5$ for the rest of experiments because of its relative efficiency.

\textbf{Teacher-Student training}. We trained our teacher models and student models using our proposed algorithm  (Figure \ref{fig:diagram}). 
The total emotion metrics for both the teachers and students in multiple generations are shown in Figure \ref{fig:total_emotion}. Our first finding is that ensembles always outperform single models on the total emotion metric. As we increased the number of generations, the performance gap between the single models and ensembles became smaller. This is probably because the variability between models becomes smaller and smaller after more and more generations of self-distillation. 

Our second finding on the results of EMENet-V is that the emotion performance does not increase monotonically as the number of generations increases. This is consistent with \cite{mobahi2020self}, where they interpreted increasing generations of self-distillation as amplifying regularization. The best number of generations for EMENet-V was three. More generations added too much regularization, leading to poorer performance on the validation set. We found the same phenomenon in the results of EMENet-VA. The best number of generations was also three.
We evaluated our models on the test set of the Aff-wild2. The results are listed in Table \ref{tab:test_res}. The visual-audio models always have better performance than visual models, although with slightly larger computation cost. 

\subsection{Uncertainty Performance}

There are two types of uncertainty we are interest in: the aleatoric uncertainty and the epistemic uncertainty. The aleatoric uncertainty arises from the natural complexities of the underlying distribution, such as
class overlap, label noise, input data noise, \etc. The epistemic uncertainty arises from a lack of knowledge about the best model parameters. It can be explained away given enough data in that region. The latter uncertainty is an indicator of out-of-distribution (OOD) samples. 

We capture the two types of uncertainty with deep ensembles. Following \cite{depeweg2018decomposition} and \cite{malinin2019ensemble}, we compute the two types of uncertainty as follows:
\begin{align}
    \underbrace{\mathcal{H}(E_{p(\theta|\mathcal{D})}\left[P(y|x, \theta)\right])}_{total\; uncertainty} = \underbrace{E_{p(\theta|\mathcal{D})}\left[\mathcal{H}(P(y|x, \theta)\right]}_{aleatoric\; uncertainty} + \nonumber \\  \underbrace{\mathcal{MI}\left[P(y), \theta|x, \mathcal{D}\right]}_{epistemic\; uncertainty} .
    \label{eq:uncertainty_decomposition}
\end{align}
$P(y)$ is the probabilistic distribution output of a single model. $\mathcal{H}$ is the Shannon's entropy. $p(\theta|\mathcal{D})$ is the posterior distribution of the model weights $\theta$ which is trained on dataset $\mathcal{D}$.  

\textbf{In-domain uncertainty}. The aleatoric uncertainty is an indicator of noise inherent in in-domain data. It is computed as the average entropy of single models' outputs in an ensemble, as shown in Equation \ref{eq:uncertainty_decomposition}. To evaluate the performance of in-domain uncertainty, we computed the average NLL (or RMSE) for single models to measure the similarity between predicted probability distributions and the true probability distributions. 

To evaluate AU uncertainty estimation performance, we computed the NLL values on the AU validation set. Besides the NLL values of our single teacher models and single student models, the NLLs values of Temperature scaling (TS) \cite{guo2017calibration} and Monte-Carol Dropout (MC Dropout) \cite{gal2016dropout} were also computed. For a fair comparison, we adopted the test-time cross-validation in \cite{ashukha2020pitfalls} to compute the NLL in TS. The optimal temperature was optimized on a randomly-split half of the validation set. The NLL was then evaluated on the other half of the validation set. For MC Dropout, we averaged the probability outputs of ten forward passes, where the model weights were sampled randomly for every forward pass from the dropout. 

Table \ref{tab:AU_uncertainty} shows the NLL values of single models, where the model architecture is EMENet-V. The validation performance for the EMENet-VA architecture is given in Appendix C. Table \ref{tab:AU_uncertainty} shows that the single student models in later generations (\ie., 2, 3 and 4) have better uncertainty estimation performance than TS and MC Dropout. When it comes to the averaged NLL values, our method outperforms TS by $10.8\%$ and MC Dropout by $9.5\%$. 

In Table \ref{tab:expr_va}, we list the uncertainty metrics for facial expressions (EXPR), valence and arousal. We find that the single student models have better uncertainty performance than TS and MC Dropout for both tasks. Our algorithm improves the EXPR NLL by $10.3\%$ when compared with TS and $15.5\%$ when compared with MC Dropout. As for valence/arousal RMSE scores, our method outperforms MC Dropout by $7.0\%$/$8.0\%$.

\textbf{Out-of-domain Uncertainty}. Epistemic uncertainty is an indicator of data insufficiency in certain regions of the input space. We expect that out-of-domain samples will be associated with large epistemic uncertainty. The epistemic uncertainty is computed by subtracting the aleatoric uncertainty from the total uncertainty. 

\begin{figure}[hbtp]
    \centering
    \includegraphics[width=\columnwidth]{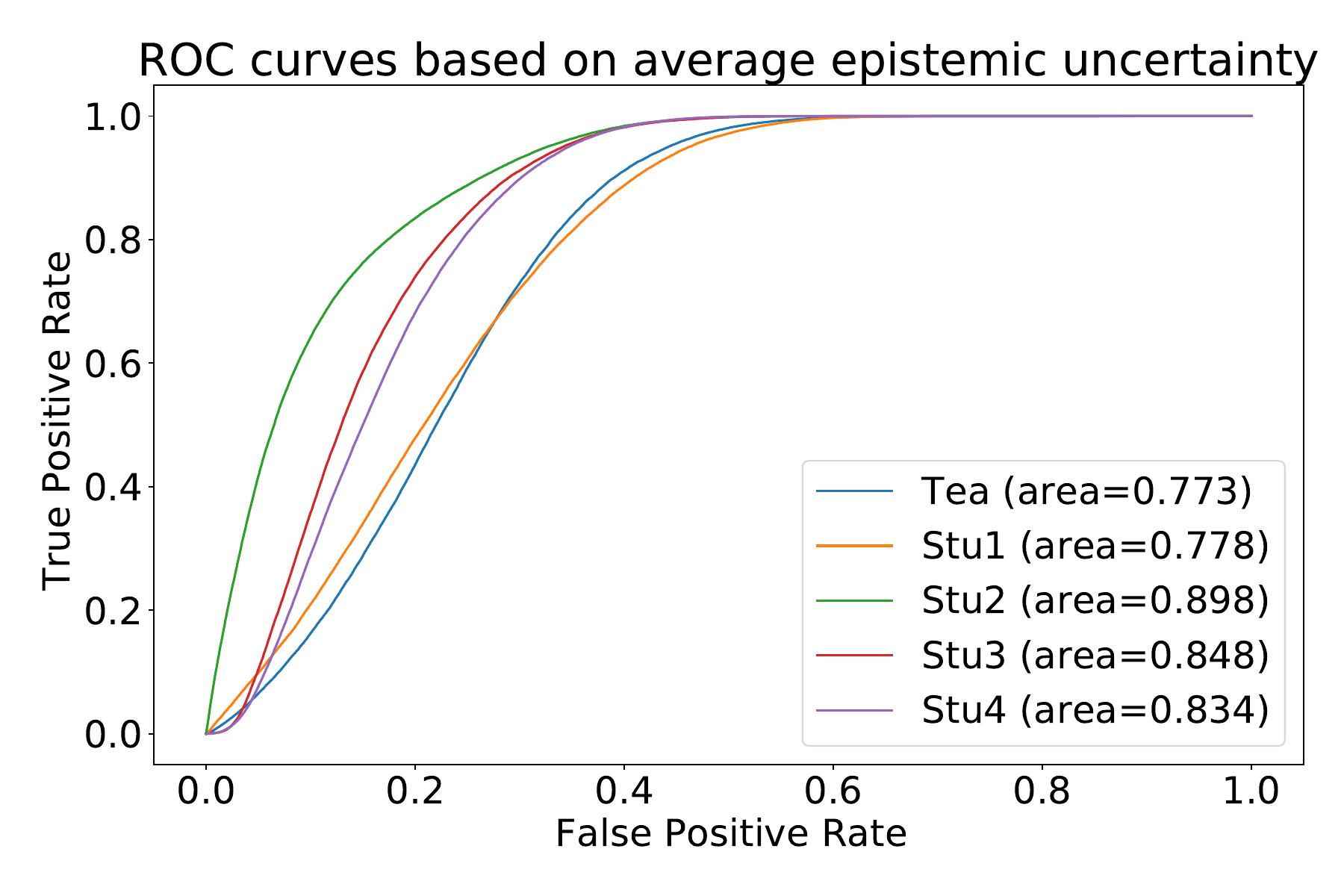}
    \caption{The ROC curves and AUC scores for the OOD detection task we created (Fashion-MNIST training set are OOD samples). Each curve corresponds to the ensemble in a certain generation. The model architecture for the results is EMENet-V.}
    \label{fig:total_ood}
\end{figure}

To evaluate the performance of the epistemic uncertainty computed by our models for detecting out-of-domain samples, we chose the Fashion-MNIST \cite{xiao2017fashion} training set as OOD samples. It contains 60,000 gray-scale images with objects like bags, shirts, trousers, \etc. The Aff-wild2 validation set has over 500,000 images. By mixing the Aff-wild2 validation set with the the Fashion-MNIST training set, we created an OOD detection task. We used the ensembles to compute the epistemic uncertainty of each sample. Since we performed multitask emotion prediction, the ensemble model produced an epistemic uncertainty value for each task. We averaged the epistemic uncertainties over all emotion tasks, and used a threshold to classify samples into in-domain and out-of-domain samples. We plotted the ROC curves showing the OOD performance across all thresholds. The ROC curves computed on the averaged epistemic uncertainty are shown in Figure \ref{fig:total_ood}. Each ROC curve corresponds to the ensemble in one generation. The AUC scores are also computed to show the OOD performance. Higher AUC scores indicate more accurate differentiation between in-domain and out-of-domain samples.

From Figure \ref{fig:total_ood}, we had the best AUC score for the student ensemble in the second generation. The AUC scores show a unimodal relationship with the number of generations. As we distill more and more, the uncertainty performance gradually increases to its best value, then decreases. 

\textbf{Summary}. We find that self-distillation improves both types of uncertainty. The epistemic uncertainty produced by our ensembles can detect out-of-domain samples accurately. The best AUC score was $0.898$. When estimating in-domain uncertainty, ensembling is not required. If the computation resources are limited, it is acceptable to use the probability outputs of a single model to compute the entropy: as an approximation to the aleatoric uncertainty computed from an ensemble's output (Equation \ref{eq:uncertainty_decomposition}).

\section{Conclusions}
In this paper, we propose to apply deep ensemble models learned by a multi-generational self-distillation algorithm to improve emotion uncertainty estimation. Our designed model architectures are efficient, and can be potentially applied in mobile devices. Our experimental results show that our algorithm can improve both the emotion metrics and uncertainty metrics as the number of generations increases. The uncertainty estimates given by our models are reliable indicators of in-domain and out-of-domain samples. In the future, we will study the regularization effect of the self-distillation algorithm, and seek better regularization methods to replace the time-consuming progress of self-distillation.
\section{Acknowledgements}
This work was supported in part by the Hong Kong Research Grants Council under grant number 16213617.

{\small
\bibliographystyle{ieee_fullname}
\bibliography{egbib}
}
\clearpage

\appendix
\section*{Appendices}
\addcontentsline{toc}{section}{Appendices}
\renewcommand{\thesubsection}{\Alph{subsection}}

\subsection{Data Distribution}
\begin{figure}[ht]
  \centering
  \begin{subfigure}[b]{0.8\columnwidth}
    \includegraphics[width=\textwidth]{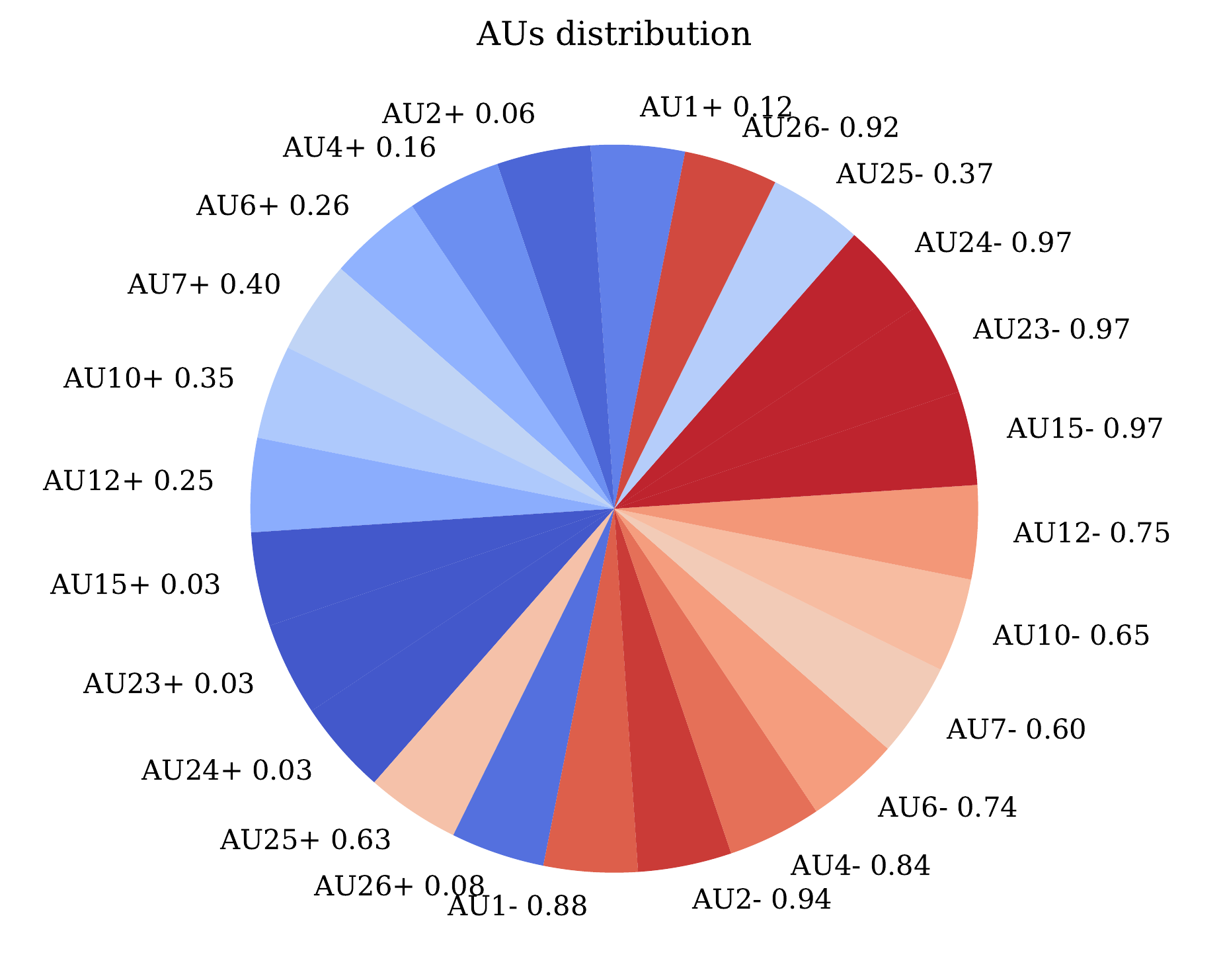}
    \caption{The data distribution for the AU subset.}
    
  \end{subfigure}
  \begin{subfigure}[b]{0.8\columnwidth}
    \includegraphics[width=\textwidth]{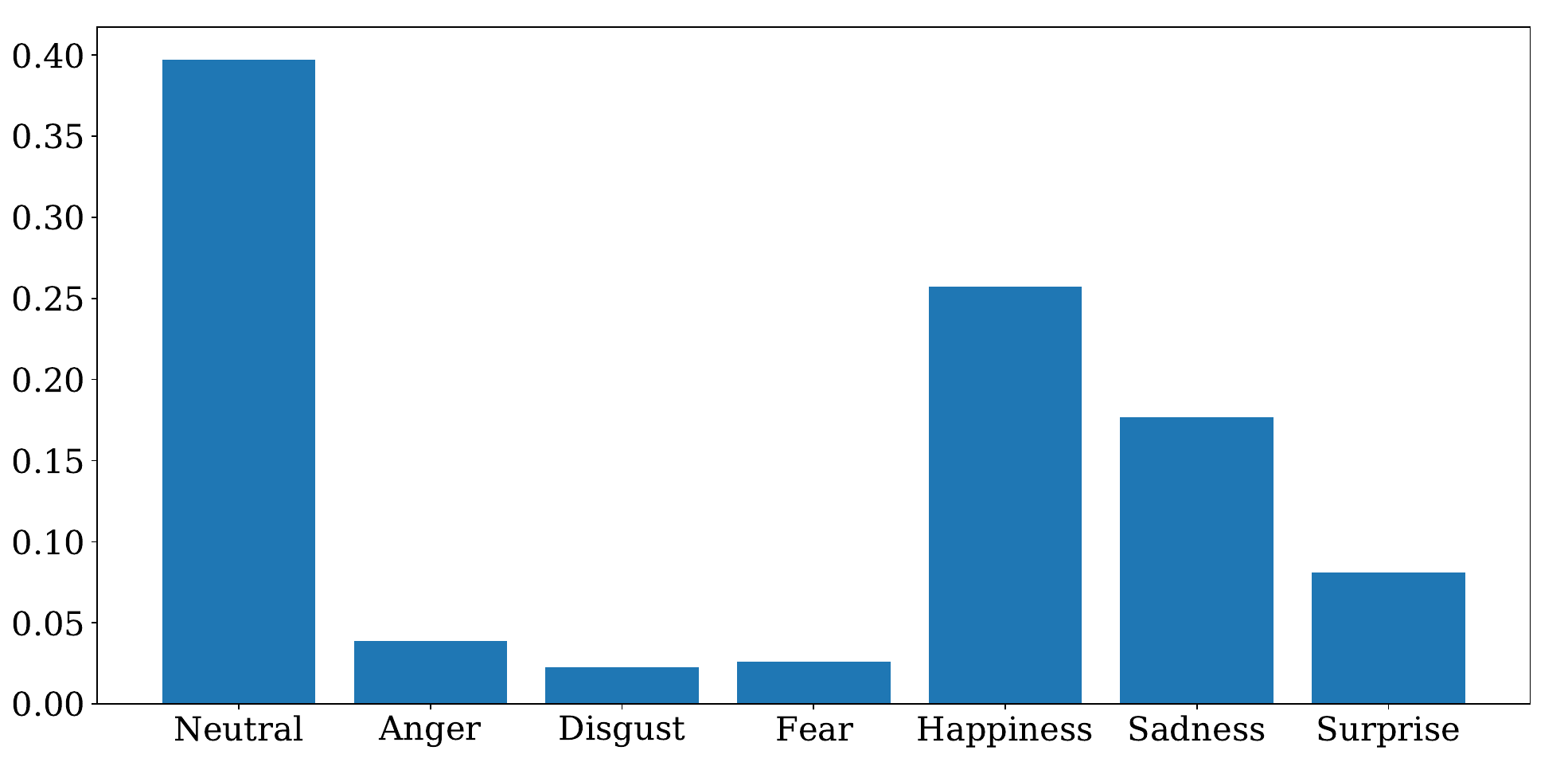}
    \caption{The data distribution for the EXPR subset.}
  \end{subfigure}
    \begin{subfigure}[b]{0.8\columnwidth}
    \includegraphics[width=\textwidth]{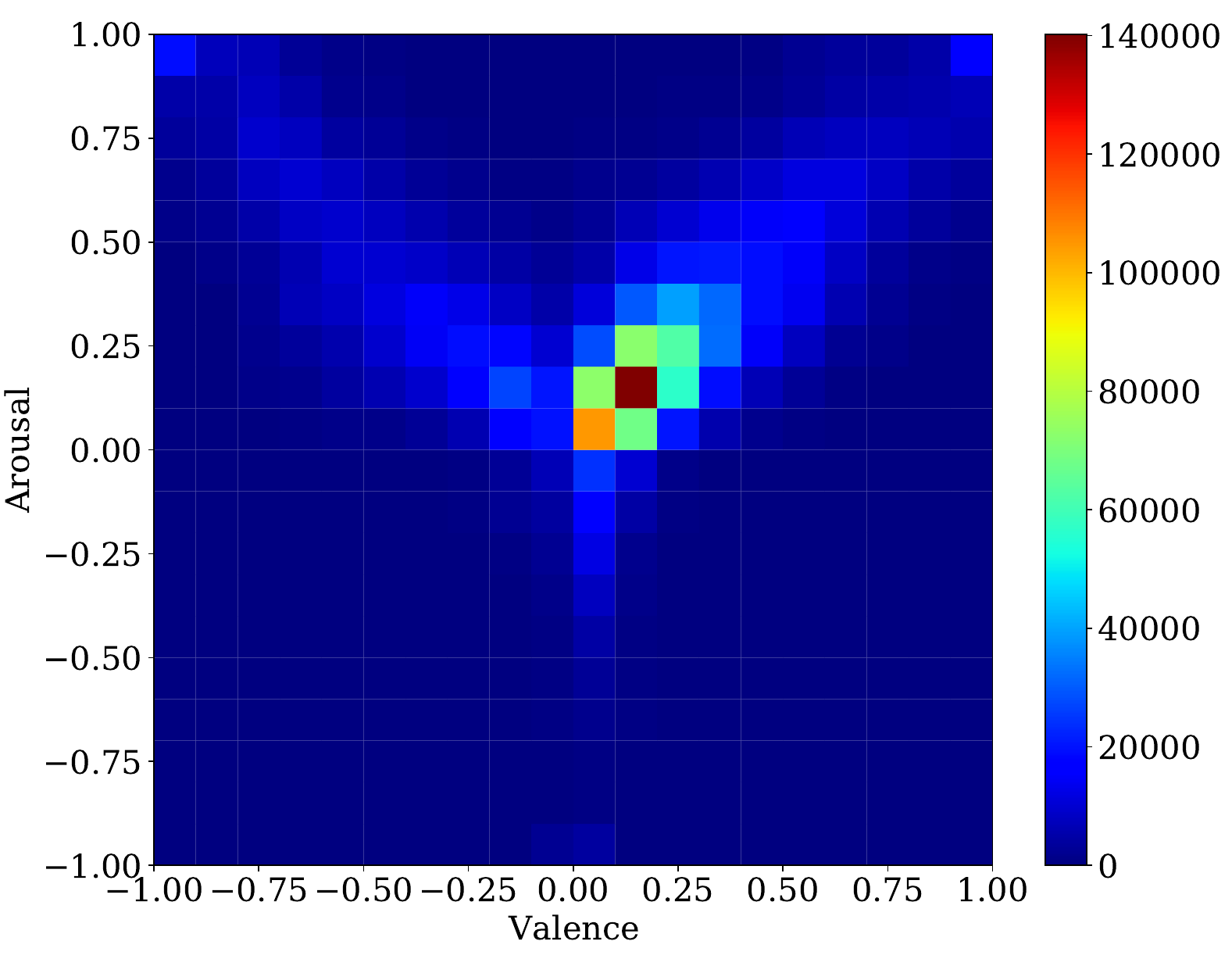}
    \caption{The data distribution for the VA subset.}
  \end{subfigure}
  \caption{The data distributions of the Aff-wild2 dataset. }
  \label{fig:data_distribution}
\end{figure}

The $p_c$ for class reweighting depends on the data distributions. From AU1 to AU26, $\mathbf{p} = [7.7, 24.7, 5.3, 2.9, 1.5, 1.9, 3, 32.3, 32.3, 32.3, 0.59, 11.5]$ to alleviate the unbalanced data problem. For the EXPR subset, $\mathbf{p} = [0.02, 0.2, 0.33, 0.24, 0.03, 0.05, 0.1]$ in Equation \ref{eq: ce_expr}.

\subsection{Multitask Balancing}

Our multitask balancing algorithm is given as follows:

    \begin{algorithm}[htbp]
    \caption{Balancing Multitask Weights}\label{euclid}
    \hspace*{\algorithmicindent} \textbf{Input} \\
    \hspace*{1em} The multitask model $f$. \\
    \hspace*{1em} The training set $\mathcal{D}_{train}$ and the validation set $\mathcal{D}_{val}$. \\
    \hspace*{1em} Tasks set $\mathcal{T}=\{AU, EXPR, VA\}$. \\
    \hspace*{1em} The number of tasks $n(\mathcal{T}) = 3$. \\ 
    \hspace*{1em} The number of epochs with no performance improvement $\mathcal{M} = \{m^{AU}, m^{EXPR},m^{VA}$. \\
    \hspace*{1em} The weights for all tasks  $\Lambda= \{\lambda^{AU}, \lambda^{EXPR},\lambda^{VA}\}$. \\
    \hspace*{1em}  The number of training epochs $M$.
    \begin{algorithmic}[1]
    \Procedure{}{}
    \While{$i < n(\mathcal{T})$}
    \State $m^{i} = 1$
    \State $\lambda^i = \frac{1}{n(\mathcal{T})}$
    \EndWhile
    
    \While{$\mathrm{i\_epoch} < M$} 
    \State Optimize $f$ on $\mathcal{D}_{train}$.
    \State Evaluate $f$ on $\mathcal{D}_{val}$.
    \While{$i < n(\mathcal{T})$}
    \State $Val^{i} \gets$ validation performance of the $i^{{th}}$ \hspace*{5em}task.
    \If{$Val^{i}$ is improved} 
    \State $m^i \gets 1$
    \Else 
    \State $m^i \gets m^i+1$
    \EndIf
    \EndWhile
    \While{$i<n(\mathcal{T})$}
    \State $\lambda^i = max(1, \log_2(m^i))$
    \EndWhile
     \While{$i<n(\mathcal{T})$}
    \State $\lambda^i = \frac{\lambda^i}{\sum_i \lambda^i}$
    \EndWhile
    \EndWhile
    \EndProcedure
    \end{algorithmic}
    
    \label{alg:balance_weights}
    \end{algorithm}

\begin{table}[hbtp]
    \centering
    \resizebox{\columnwidth}{!}{
    \begin{tabular}{c|c|c|c|c|c|c}
    \hline
        \multirow{2}{*}{Experiments} & \multirow{2}{*}{T}  & \multirow{2}{*}{AU} & \multirow{2}{*}{EXPR}  & \multicolumn{2}{c}{VA} & \multirow{2}{*}{Total Emotion} \\ \cline{5-6}
       & &   & &  Valence & Arousal  & \\
       \hline 
       w/o ba. & 1  & 0.6307 & 0.5620 & 0.3902  & 0.5344 & 2.1173\\
        w/o ba. & 5 & 0.6487 & \textbf{0.5802} & 0.4221 &\textbf{ 0.5585 }& 2.2095 \\
       w/ ba. &1 & 0.6632  & 0.5541 & 0.4202 & 0.5192 & 2.1527 \\
       w/ ba. & 5  & \textbf{0.6808} &0.5779 & \textbf{0.4423} & 0.5455 &\textbf{2.2465}\\

       \hline
        \end{tabular}
        }
    \caption{Experiment results with the teacher models using visual modality only. "w/ ba." means we apply Algorithm \ref{alg:balance_weights} for multitask training. $T$ is the number of models in an ensemble. Total emotion metric is the sum of all metrics of the three emotion tasks. }
    \label{tab:balancing_ablation} 
\end{table}
\begin{table*}[hbtp]
    \centering
     \resizebox{\textwidth}{!}{
    \begin{tabular}{c|c|c|c|c|c|c|c|c|c|c|c|c|c}
    \hline
        Methods  & AU1 & AU2 &AU4 & AU6 & AU7 & AU10 & AU12 & AU15 & AU23 & AU24 & AU25 & AU26 & Avg. \\ 
    \hline
     Tea &0.380 & 0.302 & 0.427 & 0.407 & 0.472 & 0.422 & 0.352 & 0.307 & 0.378 & 0.394 & 0.466 & 0.401 & 0.392\\
     Stu1 &0.326 & 0.264 & 0.366 & \textbf{0.381} & \textbf{0.459} & \textbf{0.412} & 0.355 & \textbf{0.256} & 0.289 & 0.256 & \textbf{0.458} & 0.359 & \textbf{0.348}\\
     Stu2 &0.320 & 0.261 & 0.368 & 0.384 & 0.465 & 0.419 & 0.366 & \textbf{0.256} & 0.280 & 0.251 & 0.467 & 0.352 & 0.349\\
     Stu3 &\textbf{0.308} & \textbf{0.255} & \textbf{0.359} & 0.388 & 0.469 & 0.427 & 0.383 & 0.262 & \textbf{0.276} & \textbf{0.233} & 0.474 & \textbf{0.342} & \textbf{0.348}\\
      TS \cite{guo2017calibration} & 0.380 & 0.301 & 0.415 & 0.406 & 0.472 & 0.421 & \textbf{0.351} & 0.306 & 0.377 & 0.381 & 0.465 & 0.400 & 0.390\\
      MC \cite{gal2016dropout}& 0.356 & 0.274 & 0.419 & 0.411 & 0.473 & 0.416 & \textbf{0.351} & 0.311 & 0.345 & 0.384 & 0.463 & 0.382 & 0.382 \\
       \hline
        \end{tabular}
        }
    \caption{The NLL values for 12 action units, which are evaluated on the validation set of the Aff-wild set. We compare our single teacher models and single student models with other methods, \ie, TS (temperature scaling \cite{guo2017calibration}) and MC (Monte-Carol Dropout \cite{gal2016dropout}). The model architecture used in this comparison is EMENet-VA.} 
    \label{tab:AU_uncertainty_EMENet-VA} 
\end{table*}

\begin{table}[hbtp]
    \centering
    \begin{tabular}{c|c|c|c}
    \hline
        \multirow{2}{*}{Methods} & \makecell{EXPR \\ NLL}& \makecell{Valence \\ RMSE} &\makecell{Arousal \\ RMSE}   \\
        \hline
        Tea & 1.060 & 0.416 & 0.240 \\
        Stu1 & \textbf{0.825} & 0.397 & 0.233 \\
        Stu2 & 0.846 & 0.392 & 0.231\\
        Stu3 & 0.858 & \textbf{0.383} & \textbf{0.230}\\
        TS \cite{guo2017calibration} & 0.955 & - & -\\
        MC \cite{gal2016dropout} & 0.994 & 0.416 & 0.237 \\
        \hline
    \end{tabular}
    \caption{The NLL values for EXPR recognition and the RMSE values for valence and arousal prediction. Metrics are evaluated on the validation set. TS optimizes temperature for lower NLL on a held-out validation set, which is not beneficial for RMSE in regression tasks. Therefore, we only compare our models with TS for EXPR task. }
    \label{tab:expr_va_EMENet-VA}
\end{table}

The main idea of this algorithm is to increase the weight of certain task if this task has not been improved on the validation set for a number of epochs. Once this task has been improved, the weight of the loss function for this task is set to its initial value. 

We conducted ablation studies on the effect of Algorithm \ref{alg:balance_weights}. The model we used is the EMENet-V trained on original dataset. From the experiment results in Table \ref{tab:balancing_ablation}, we notice that the EXPR and arousal metrics are better without multitask balancing algorithm, but the total emotion metric is better when using multitask balancing algorithm. We value the performance of each emotion tasks equally. The Algorithm \ref{alg:balance_weights} was used in all other experiments for better total emotion metric.

\subsection{In-domain Uncertainty for EMENet-VA}

We show the AU uncertainty performance using the EMENet-VA in Table \ref{tab:AU_uncertainty_EMENet-VA}. We also compare our single teacher models and single student models with Temerature Scaling (TS) and Monte-Carol (MC) Dropout. Similar to the results in Table \ref{tab:AU_uncertainty} (EMENet-V), the models using our algorithm achieved the lowest NLL value, compared with TS and MC Dropout. The lowest AVg. NLL value is 0.348 for EMENet-VA, while for EMENet-V, the lowest average NLL value is 0.381. We find that when using audio features with visual features, the uncertainty (NLL) of facial actions can be improved by about $8.7\%$.

Table \ref{tab:expr_va_EMENet-VA} shows the uncertainty performance for the EXPR task, valence and arousal detection. We evaluated NLL for classification tasks and RMSE for regression tasks. Comparing Table \ref{tab:expr_va_EMENet-VA} with Table \ref{tab:expr_va}, we find that incorporating audio features with visual features, it improved the EXPR NLL by $5.2\%$. However, it failed to improve the valence RMSE, and barely had an influence on the arousal RMSE.

\end{document}